\documentclass[runningheads]{llncs}

\usepackage[mobile]{eccv}

\usepackage{eccvabbrv}

\usepackage{graphicx}
\usepackage{booktabs}

\usepackage[accsupp]{axessibility}  
\usepackage{float}
\usepackage{pifont}
\usepackage[section]{placeins}
\usepackage{colortbl}
\usepackage{multicol}
\usepackage{caption}

\usepackage{hyperref}

\usepackage{orcidlink}

\begin{document}

\title{\footnotesize This work has been submitted to the IEEE for possible publication. Copyright may be transferred without notice, after which this version may no longer be accessible.  \\
\vspace{20pt}
\large Graph and Skipped Transformer: Exploiting Spatial and Temporal Modeling Capacities for Efficient 3D Human Pose Estimation}

\titlerunning{Abbreviated paper title}

\author{Mengmeng Cui\inst{1}\orcidlink{0000-0003-4281-3125} \and
Kunbo Zhang\inst{1}\orcidlink{0000-0002-4826-6831} \and
Zhenan Sun\inst{1}}

\authorrunning{M. Cui et al.}

\institute{
New Laboratory of Pattern Recognition (NLPR), Institute of Automation, Chinese Academy of Sciences, Beijing 100190, China\\
\email{mengmeng.cui@cripac.ia.ac.cn} }

\maketitle
\begin{abstract}

In recent years, 2D-to-3D pose uplifting in monocular 3D Human Pose Estimation (HPE) has attracted widespread research interest. GNN-based methods and Transformer-based methods have become mainstream architectures due to their advanced spatial and temporal feature learning capacities. However, existing approaches typically construct joint-wise and frame-wise attention alignments in spatial and temporal domains, resulting in dense connections that introduce considerable local redundancy and computational overhead. In this paper, we take a global approach to exploit spatio-temporal information and realise efficient 3D HPE with a concise Graph and Skipped Transformer architecture. Specifically, in Spatial Encoding stage, coarse-grained body parts are deployed to construct Spatial Graph Network with a fully data-driven adaptive topology, ensuring model flexibility and generalizability across various poses. In Temporal Encoding and Decoding stages, a simple yet effective Skipped Transformer is proposed to capture long-range temporal dependencies and implement hierarchical feature aggregation. A straightforward Data Rolling strategy is also developed to introduce dynamic information into 2D pose sequence. Extensive experiments are conducted on Human3.6M, MPI-INF-3DHP and Human-Eva benchmarks. G-SFormer series methods achieve superior performances compared with previous state-of-the-arts with only around ten percent of parameters and significantly reduced computational complexity. Additionally, G-SFormer also exhibits outstanding robustness to inaccuracies in detected 2D poses.

\end{abstract}

\section{Introduction}
\label{sec:intro}

3D Human Pose Estimation (HPE) is a fundamental task which aims to reconstruct 3D body joint locations from images or videos. Monocular 3D HPE is more friendly for downstream applications such as action recognition \cite{shi2019two,yan2018spatial,shi2019skeleton}, human-computer interaction \cite{sinha2010human,lazar2017research,pavlovic1997visual}, motion and trajectory prediction \cite{martinez2017human,wang2021multi,rudenko2020human} for the convenience in data acquisition. 

Benefits from rapid development in 2D pose detectors \cite{chen2018cascaded,sun2019deep,cao2017realtime,rogez2019lcr,he2017mask}, 2D-to-3D pose lifting methods have drawn extensive attentions for its high spatial precision and light data volume of 2D skeletons. Despite the superior performance, 2D-to-3D lifting method is inherently an ill-pose problem for depth ambiguity and self-occlusion \cite{cheng2019occlusion,cheng20203d,li2022mhformer}. To alleviate this issue, many recent works incorporate temporal correlations in videos to pose reconstruction process. Mainstream methods can be devided into CNN-based \cite{pavllo20193d, chen2021anatomy,zhou2017towards,liu2020attention}, GCN-based \cite{wang2020motion,cai2019exploiting,hu2021conditional,yu2023gla} and Transformer-based \cite{zheng20213d,li2022mhformer,li2022exploiting, shan2022p,einfalt2023uplift}. With the extension of temporal receptive field, long-range dependencies are established and performance is further improved. Meanwhile, taking into account practical application requirements, 3D HPE models are more inclined to have low computation budgets for the deployment on resource-limited mobile devices and consumer hardware.

Due to the natural correlation with human skeleton structure, Graph Neural Networks (GNN) are widely adopted to represent human pose in deep-learning-based computer vision tasks, e.g., action recognition \cite{shi2019two,yan2018spatial}, pose estimation \cite{wang2020motion, cai2019exploiting, hu2021conditional}. Since Yan et al. \cite{yan2018spatial} firstly introduced GCN to model human skeletons, many researches have been proposed to develop more adaptive graphs. Shi et al. \cite{shi2019skeleton} fix graph topology with predefined connections in early training stage and release it afterwards. Hu et al. \cite{hu2021conditional} calculate conditioned connections based on predefined connections and combine them both for graph construction. In 2s-AGCN \cite{shi2019two} and GLA-GCN \cite{yu2023gla}, a parameterized graph and a data-dependent graph are added to the predefined graph structure. However, the predefined graph based on physical body structure is used in all these methods to maintain performances and stabilize training, which somewhat restricts model flexibility. Meanwhile, building join-wise connections in graphs also introduces computational redundancy, as only parts of human body need to cooperate for specific actions. For example, two arms are closely related during "Eating", while two legs are correlated for "Sitting". To address these problems, we deploy coarse-grained body parts to construct a compact graph structure, of which the topology is adaptively learned through graph attention mechanism. The completely data-driven approach ensures model flexibility and increases generality to various poses.

\begin{figure}[h]
    \centering
    \includegraphics[width=0.9\textwidth]{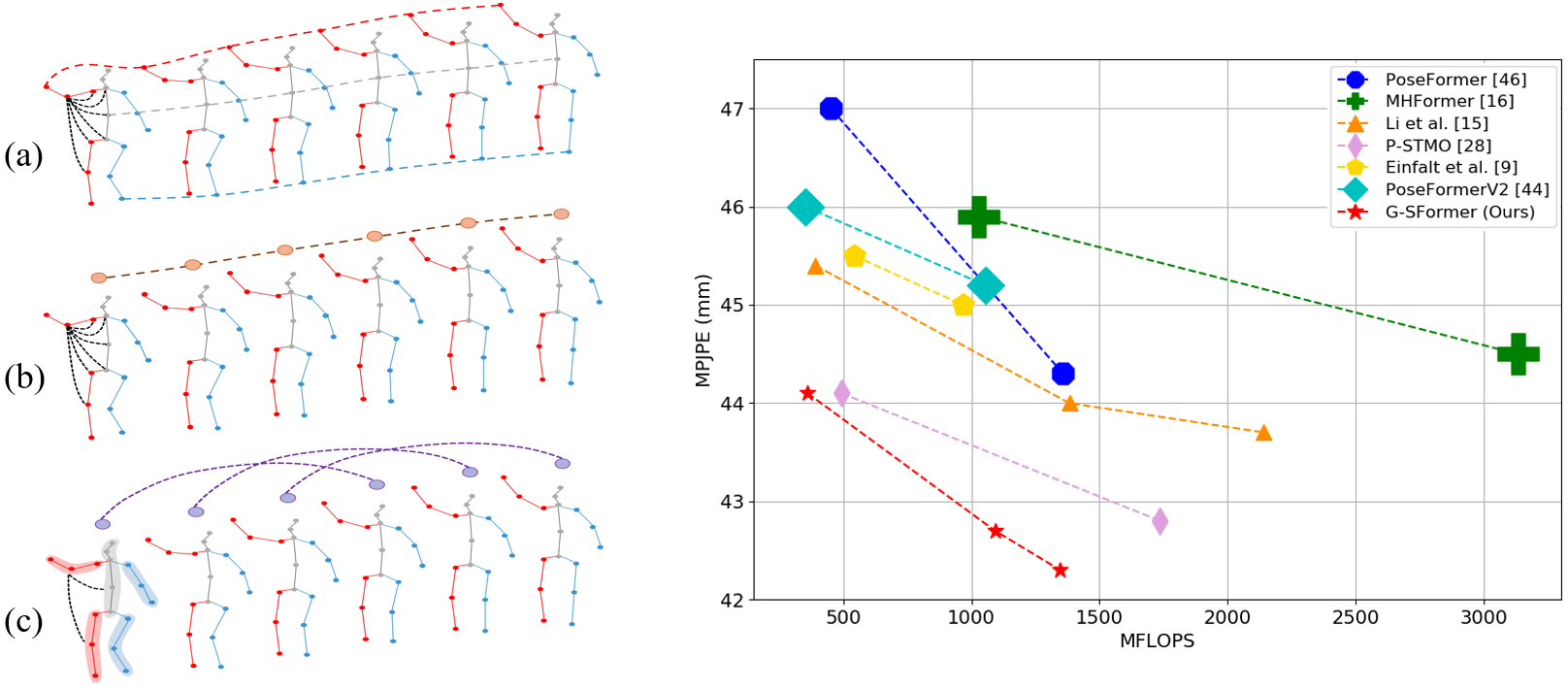}
    \caption{\textbf{Left}: Comparison of spatio-temporal correlation modeling methods: (a) building joint-wise connections and frame-wise connections for each joint (b) building joint-wise connections and frame-wise connections for the pose representation (c) our G-SFomer: constructing part-based spatial alignments and long-range temporal skipped-connections. \textbf{Right}: MPJPE (mm) vs. MFLOPs of the proposed G-SFormer and competitors on Human3.6M dataset, where marker size indicates model size.}
    \label{fig0}
\end{figure}

Transformer architecture has become prevalent in recent 2D-to-3D HPE approaches for its long-range dependency modeling capacity. Prior methods typically deploy self-attention to establish joint-wise correlations, as well as frame-wise correlations for each joint individually \cite{zhang2022mixste}  or for the encoded pose representation \cite{zheng20213d, einfalt2023uplift} (as shown in Figure \ref{fig0} (a) and (b)). This is computationally expensive especially when dealing with lengthy sequences (81, 243 or even more). Meanwhile, information redundancy in adjacent frames also makes improving the efficiency of Transformer an urgent problem to be solved. Some efforts have been made on this direction, e.g., Li et al. \cite{li2022exploiting} and Shan el al. \cite{shan2022p} introduce strided convolutional layers to FFN in transformer to selectively aggregate useful information; Einfalt et al. \cite{einfalt2023uplift} perform upsampling on spatially encoded pose tokens for temporal uplifting; Zhao et al. \cite{zhao2023poseformerv2} fuse selected tokens in the time domain and the frequency domain to form a compact pose representation. However, it is worth noting that reducing pose tokens in temporal domain will lead to performance degradation since partial information is inevitably lost. Furthermore, none of them cut to the biggest computational overhead -- the Self Attention calculation which is quadratic to the number of tokens. To this end, a Skipped Transformer with Skipped Self-Attention (SSA) mechanism is proposed to construct long-term temporal correlations. SSA is deployed in both Temporal Encoder and Decoder for hierarchical feature extraction and aggregation. This approach establishes alignments among distinct frame tokens, reducing computation redundancy without sacrificing information integrity.

In this paper, we propose a novel Graph and Skipped Transformer (G-SFormer) architecture. Different from existing Graph-Transformer hybrid methods \cite{zhu2021posegtac,zhao2022graformer} that embed GCN into Transformer block to assist self-attention in spatial modeling, G-SFormer integrates part-based adaptive GNN and Skipped Transformer to efficiently exploit spatial and temporal information, respectively. The compact and adaptive framework significantly reduces redundant connections existing in previous GCN-based and Transformer-based methods, enabling efficient and robust 3D pose reconstruction in a global approach. Extensive experiments are conducted on three benchmarks, i.e., Human3.6M, MPI-INF-3DHP, and HumanEva. G-SFormer achieves sate-of-the-art performances and generalizes well to large and small datasets without any pre-training process, which is frequently adopted in transformer-based counterparts \cite{zheng20213d,zhang2022mixste, shan2022p,einfalt2023uplift}. More importantly, as shown in Figure \ref{fig0}, the proposed G-SFormer outperforms previous state-of-the-art methods in accuracy and speed. Our main contributions are summarized as follows:

\begin{itemize}
\item

We propose a novel G-SFormer architecture, incorporating Part-based Adaptive GNN and Skipped Transformer to perform efficient feature extraction and aggregation in spatial and temporal domains. 





\item
A new Skipped Self-Attention mechanism is introduced into Transformer to establish informative global-range dependency. Both theoretical analysis and experimental results demonstrate a significant reduction in computational complexity by nearly 50\% without compromising accuracy. Additionally, by integrating an adaptive GNN for part-based spatial modeling, G-SFormer achieves superior performance among various datasets with significantly fewer parameters and lower computational cost.


\item
We also propose data completion methods to address data missing problems by introducing rich dynamic information into the input 2D pose sequence.
\end{itemize}

\vspace{-0.5cm}
\section{Related Works}
\vspace{-0.2cm}
Mainstream 3D HPE approaches can be divided into two categories according to the difference in training pipelines, i.e., 1) one-stage manner by directly reconstructing 3D pose from RGB images \cite{mehta2017monocular,ma2021context,pavlakos2017coarse,tekin2016direct}. 2) two-stage manner by obtaining 2D poses with off-the-shelf pose detectors \cite{chen2018cascaded,sun2019deep}, and then uplifting 2D poses to 3D body joint locations \cite{liu2020comprehensive,hu2021conditional,zhao2019semantic,xu2021graph,martinez2017simple,ci2019optimizing}. The two-stage methods taking 2D pose sequence as input are now dominant in 3D HPE for the ability to alleviate depth ambiguity by exploiting temporal information. Meanwhile, compact 2D poses are more memory friendly compared with image-pose pairs employed in one-stage approaches. The proposed G-SFormer also follows the 2D-to-3D pose lifting pipeline, using detected 2D pose sequence to estimate 3D joint locations.

\subsection{2D-to-3D Pose Lifting}
Recent lifting-based approaches exploit contextual information from neighboring frames to improve robustness and accuracy. Pavllo et al. \cite{pavllo20193d} present a fully-convolutional model with dilated temporal convolutions to regress temporal information. Chen et al. \cite{chen2021anatomy} leverage CNN-based bone length and bone direction prediction networks to derive 3D joint locations. Due to natural correlations with human skeleton structures, GCN-based methods are adopted to incorporate spatial priors. Cai et al. \cite{cai2019exploiting} construct a spatial-temporal graph to process and consolidate pose features across scales. Wang et al. \cite{wang2020motion} design a U-shaped GCN to capture motion information and leverage motion supervision for 3D sequence reconstruction. Zeng et al. \cite{zeng2021learning} propose hierarchical channel-squeezing fusion layers to extract signals and suppress noise in message passing over GCN. DiffPose \cite{holmquist2023diffpose} incorporates HPE model \cite{zhao2022graformer} into diffusion framework. Notably, it requires heatmaps in addition to 2D pose-sequence to guide 3D pose generation.

\subsection{Transformer-based 3D HPE}
Transformer architecture proposed in \cite{vaswani2017attention} has shown promising performance in computer vision \cite{zhu2020deformable, dosovitskiy2020image}. With the outstanding ability in capturing long-range dependencies, transformer with powerful self-attention mechanism is naturally introduced to 3D HPE task. Zheng et al. \cite{zheng20213d} present PoseFormer to model joint-wise relations and temporal correlations cross frames. MixSTE \cite{zhang2022mixste} captures temporal motion of each joint and stacks spatial and temporal transformer blocks several loops to strength sequence coherence. Li et al. \cite{li2022exploiting} and Shan et al. \cite{shan2022p} incorporate strided convolution into Vanilla Transformer for temporal information aggregation. Considering the characteristics of transformer, pre-training operations such as self-supervised \cite{shan2022p} and large-scale dataset based \cite{einfalt2023uplift} are implemented for performance improvement. Recent studies also concentrate on reducing the high computational complexity caused by lengthy sequence input, and provide solutions in temporal domain \cite{einfalt2023uplift} and frequency domain \cite{zhao2023poseformerv2}.

Different from previous methods which deploy transformer for both spatial and temporal modeling \cite{zhang2022mixste,zheng20213d,li2022mhformer,li2022exploiting}, we develop Part-based Adaptive GNN and Skipped Transformer to efficiently exploit the graph-structured pose data within each frame and capture long-term dependencies across entire sequence. The two modules are integrated within a compact G-SFormer architecture to achieve comprehensive performance improvements in 3D HPE.

\begin{figure}[t]
    \centering
    \includegraphics[width=0.85\textwidth]{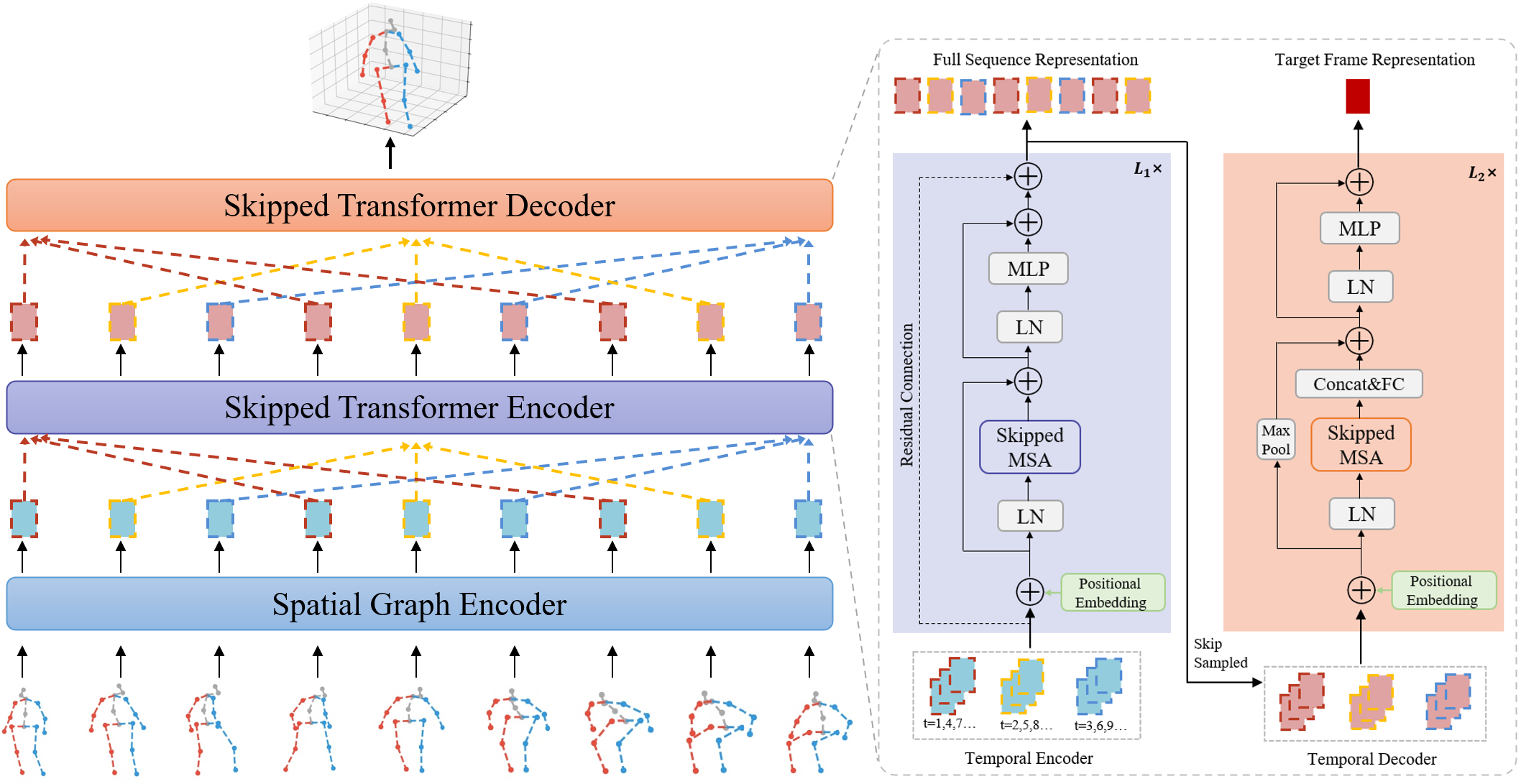}
    \caption{Graph and Skipped Transformer (G-SFormer) consists of three modules: Spatial Graph Encoder for spatial modeling of human body part correlations, Skipped Transformer Encoder and Decoder for temporal feature hierarchical extraction and aggregation. Skip-sampled pose token sets are reordered to the original sequence after encoded by temporal Skipped Transformer, and progressively aggregated by Skipped Multi-head Self-Attention (MSA) to get the target pose representation in the temporal decoding stage.}
    \label{fig01}
\vspace{-0.5cm}
\end{figure}

\vspace{-0.5cm}
\section{Method}

Following 2D-to-3D pose lifting pipelines \cite{zheng20213d,shan2022p,zhao2023poseformerv2}, the proposed G-SFormer regress 3D pose of the center frame from input 2D pose sequence estimated by off-the-shelf 2D pose detectors. As shown in Figure \ref{fig01}, Graph Neural Network (GNN) and Skipped Transformer are engaged as key components for spatio-temporal feature modeling.
\vspace{-0.2cm}
\subsection{Spatial Graph Construction}

\begin{figure}[h]
    \centering
    \includegraphics[width=0.80\textwidth]{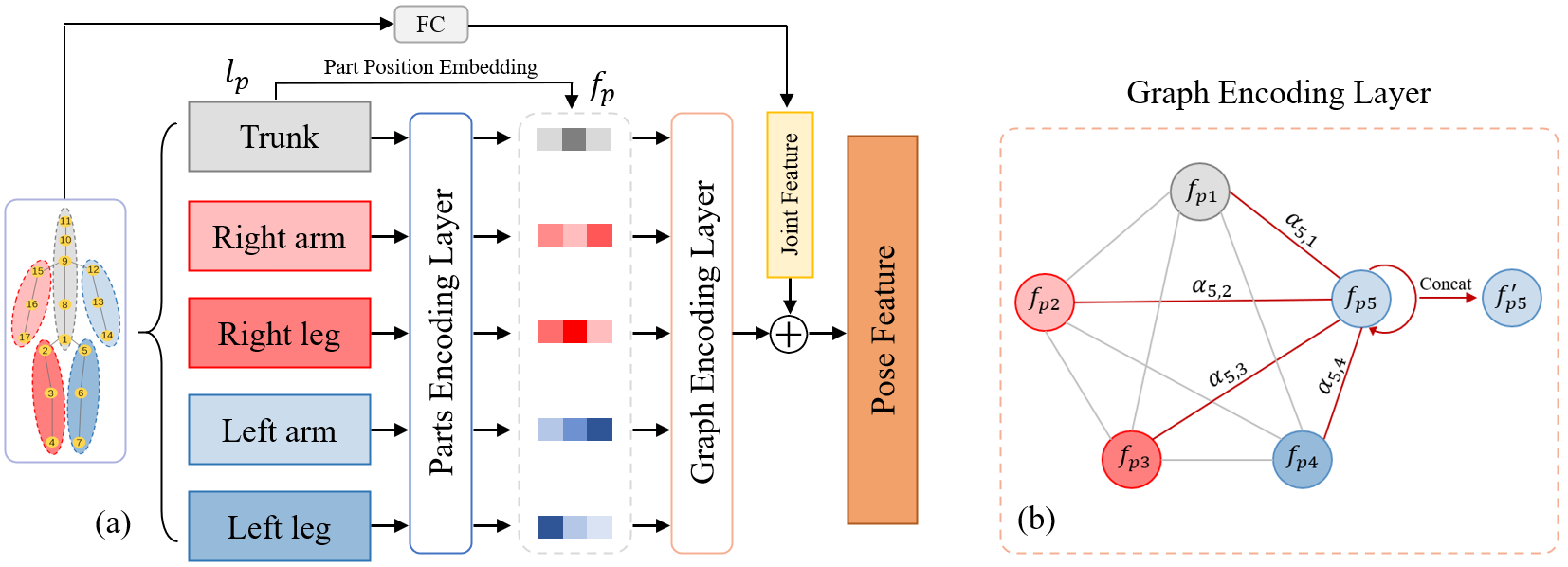}
    \caption{(a) Architecture of the Spatial Graph Encoder. (b) Updating process of graph nodes. Using node $f_{p5}$ as an example, it is concatenated with other aggregated part features to get $f_{p5}^{'}$.}
    \label{fig2}
\end{figure}

In the spatial modeling of the 2D pose in each frame, human joints are grouped into 5 parts according to their physical relationships. Trunk and limbs are used to construct graph topology of human body. We do not introduce the predefined graph structure as priors \cite{hu2021conditional,shi2019two,yu2023gla}, but adopt a data-driven approach to build a flexible graph topology to better represent the relationships between human body parts. As shown in Figure \ref{fig2} (a), the grouped 2D joint coordinates $l_{p}$ are passed into a parameter-sharing Parts Encoding Layer composed of Multi-Layer Perceptron (MLP) structure to get part feature $f_{p}$, which is a $T\times N_{p}\times C$ tensor, where $C$ is the channel dimension, $T$ is the frame length, and $N_{p}$ = 5 denotes the number of body parts. $f_{p}$ is also added with the positional embedding $E_{pos}\in \mathbb{R}^{N_{p}\times C}$ to encode the spatial position information. 
\begin{equation}
\label{eq0}
f_{p} = \textit{MLP}(l_{p}) + E_{pos} 
\end{equation}
Then, the $N_{p}\times N_{p}$ adjacency matrix is obtained by performing attention calculation among part features, where the attention coefficients can be calculated by:
\begin{equation}
\label{eq1}
e_{i,j} = \textbf{W}\times \left| f_{pi}+f_{pj} \right|
\end{equation}
Part features are added to establish alignments with each other. $\textbf{W}$ is the weight vector of $1\times 1$ convolution which transforms the dimension of attention map into 1. Attention coefficients are normalized using the $Sigmoid$ function to obtain the inter-part correlation strength.
\begin{equation}
\label{eq2}
\alpha _{i,j}=Sigmoid(e_{i,j})=1/(1+exp(e_{i,j}))
\end{equation}
Figure \ref{fig2} (b) illustrates the updating process of graph nodes. Part features are aggregated with attention coefficients to obtain the graph feature, which is then concatenated with the original part feature to get the refreshed $f_{p}^{'}$. 
\begin{equation}
\label{eq3}
f_{pi}^{'}=\bigg |\bigg | \left ( \sigma (\sum\nolimits_{j=1}^{N_{p}}\alpha _{i,j}f_{pj}),f_{pi} \right )
\end{equation}
$\sigma$ is the nonlinear activation function, and $||$ represents concatenation. In order to preserve the fine-grained spatial information, joint feature encoded by Fully Connected layer is residually added to $f_{p}^{'}$ to obtain the comprehensive pose representation. 

\subsection{Skipped Transformer for Temporal Modeling}
Skipped Transformer-based Encoder and Decoder are deployed in the temporal modeling process. Both of them work in a computation efficient manner for hierarchical temporal feature extraction and aggregation. Corresponding complexity analysis of these two stages is also presented with compression ratios of computational complexity.


\subsubsection{Skipped Transformer for Temporal Encoding}

After encoded by the spatial graph module, the spatial feature sequence is with the shape of $T \times D$. Unlike conventional self-attention mechanism that build frame-wise connections with high computational redundancy, the proposed Skipped Self-Attention (SSA) models global dependencies among distinct token representations. Specifically, skipped sampling with interval $m$ is performed on temporal dimension to construct long-range attention alignments. The sampling process is conducted $m$ times until all tokens are established associations. 
\vspace{-8pt}
\begin{equation}
\begin{split}
\label{eq4}
Attn(Q_{i}, K_{i}, V_{i}) &= Softmax(\frac{Q_{i}K_{i}^{T}}{\sqrt{D}})V_{i} \\
                          &= A_{i}V_{i},  i=1,2,...m
\end{split}
\end{equation}
where $Q_{i}$, $K_{i}$ and $V_{i}$ are obtained by linearly transformation of skipped sampling set $Z_{i}\in \mathbb{R}^{\frac{T}{m} \times D}$ by parameters $W_{Q}, W_{K}$ and $W_{V}$. 
\vspace{-8pt}
\begin{equation}
Q_{i}=Z_{i}W_{Q},  K_{i}=Z_{i}W_{K},  V_{i}=Z_{i}W_{V}
\end{equation}

The sampling set consists of $\frac{T}{m}$ tokens and the attention map $A_{i}$ is with the shape of $(\frac{T}{m})\times(\frac{T}{m})$. Therefore, the computational complexity for attention map calculation and token association is significantly reduced to:
\vspace{-4pt}
\begin{equation}
\label{eq5}
\Omega (SSA)=\frac{2T^{2}D}{m}
\end{equation}
which is \textbf{$\frac{1}{m}$} times compared with the Self-Attention in Vanilla Transformer. Thus the compression ratio can be derived as 30\%, as skipped factor $m$ is set to 3 in the proposed architecture.

Finally, the encoded tokens in each sampling set are reordered to original sequence to keep temporal dimension unchanged. Similar to \cite{vaswani2017attention}, the multi-head setting for self-attention and the feed-forward network is also deployed in the Skipped Transformer block. The temporal encoder is composed of a stack of $L_{1}$ layers for hierarchical temporal feature extraction.

\begin{figure}[h]
    \centering
    \includegraphics[width=0.8\textwidth]{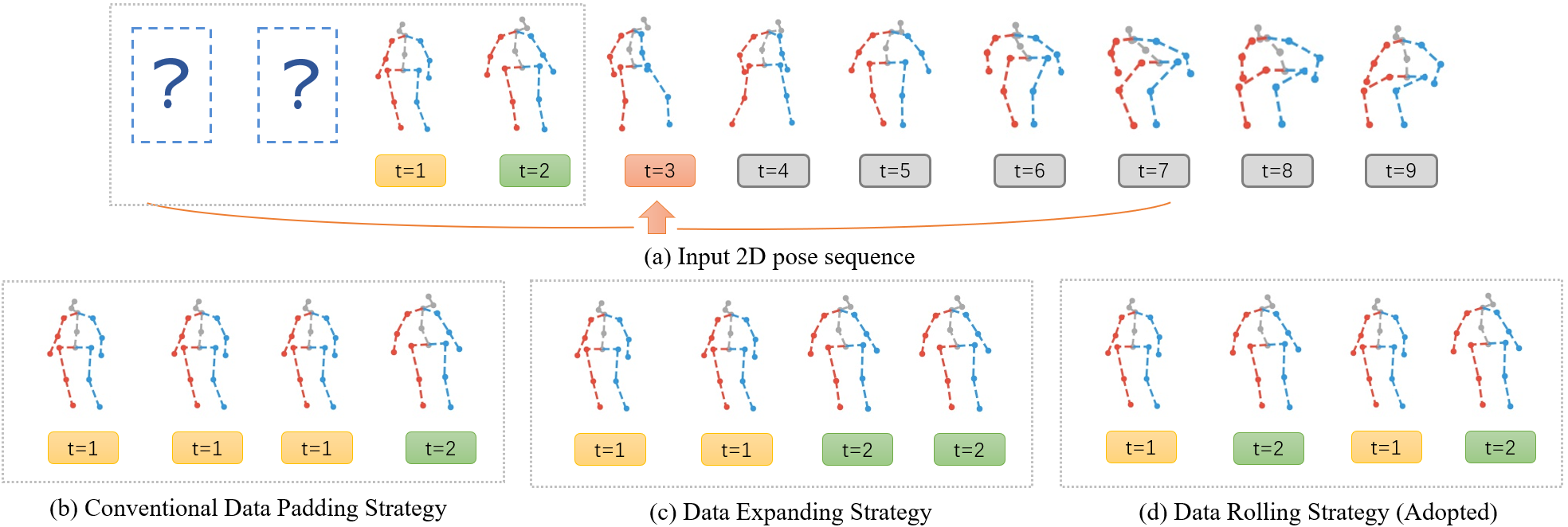}
    \caption{Data completion strategies for 2D pose input. Taking target frame at t=3 as example, where 2 previous frames need to be completed for a full 9-frame input sequence. Unlike conventional methods which copy edge frame at t=1 multiple times (b), Data Expanding and Data Rolling strategies are proposed to replicate 2D pose step by step (c), or to capture a clip of the pose sequence for completion (d).}
    \label{fig3}
\end{figure}

\subsubsection{Skipped Transformer for Temporal Decoding} 
In the decoding stage, the scale of the temporal dimension is reduced layer by layer using Skipped Self-Attention (SSA) to obtain the pose representation of target frame. Different from the encoding stage, the tokens of $m$ skipped sampling set are concatenated in the channel dimension, generating a decoded feature sequence with the shape of $\frac{T}{m} \times (m\cdot D)$, which is then transformed to $\frac{T}{m} \times D$ by a linear layer. Thus, the temporal dimension is progressively reduced by $\frac{1}{m}$ through each decoding layer until the center frame representation is obtained. The temporal decoder is composed of a stack of $L_{2}$ layers for temporal feature aggregation. 

Compared with the widely-adopted Strided Transformer \cite{einfalt2023uplift,li2022exploiting,shan2022p}, the proposed Skipped Transformer aggregates temporal features directly in self-attention calculation process, which is a simple yet effective way that reduces computational complexity significantly. For Skipped Transformer block (SKT), the complexity for key components is:
\begin{equation}
\begin{split}
\label{eq6}
\Omega (SKT)&=\frac{2T^{2}D}{m} + m(\frac{T}{m})D^{2} + 4(\frac{T}{m})D^{2} \\
            &= \frac{2T^{2}D}{m} + (1+\frac{4}{m})TD^{2}
\end{split}
\end{equation}

Where $\frac{2T^{2}D}{m}$ is for self-attention as equation \ref{eq5}, $m(\frac{T}{m})D^{2}$ is for linear transformation, and $4(\frac{T}{m})D^{2}$ is for feed-forward network (FFN). While for Strided Transformer block (STT) which uses strided convolution layer with the kernel size $k$ and strided factor $s$ to shrink sequence length, the complexity for self-attention and strided convolutional FFN in \cite{li2022exploiting,shan2022p,einfalt2023uplift} is:
\vspace{-0.2cm}
\begin{equation}
\label{eq7}
\Omega (STT)=2T^{2}D + 2(\frac{k}{s}+1)TD^{2}
\end{equation}

Quantitatively, comparing a Skipped Transformer block with $m$=3 and a Strided Transformer block with $k$=3 and $s$=3, the computational cost of the former is less than 60\% of the latter under the same setting.


\subsection{Data Completion Strategies for 2D Pose Input}
To reconstruct 3D pose of the center frame, it requires 2D poses from (T-1)/2 previous and subsequent frames to form an input sequence with length T. Hence, target frames at the beginning and end of the video suffer from missing 2D poses. The previous method is to replicate frames at $t=1$ or $t=T$ like edge padding \cite{zheng20213d,shan2022p}. However, we argue that such monotonous information has a limited help to 3D pose reconstruction. As shown in Figure \ref{fig3}, Data Expanding and Data Rolling strategies are proposed to introduce richer information for data completion. Data Expanding is to replicate 2D pose step by step, while Data Rolling is to capture a clip of pose sequence to form a scrolling input.

A threshold $R$ is set for Data Rolling strategy. Specifically, when the length of 2D poses to be completed exceeds $R$, the data completion operation will be executed. Performance comparison of Data Rolling with various $R$ and Data Expanding is provided in Section 4.4. 
\vspace{-0.3cm}
\subsection{Loss Function}
Regression heads consisting of linear transformation layers are deployed to transform full-sequence representation of Temporal Encoder $Z_{EN}\in \mathbb{R}^{T\times D}$ and target frame representation of Temporal Decoder $Z_{DE}\in \mathbb{R}^{1\times D}$ into corresponding 3D body joint locations. We use L2 loss to conduct full-sequence supervision of encoder outputs, as well as the target-frame supervision of decoder outputs. The full-to-single supervision strategy \cite{li2022exploiting, zheng20213d} benefits the optimization process and introduces temporal consistency to feature learning. Thus, the architecture is trained in an end-to-end manner with the objective loss function as:
\vspace{-0.1cm}
\begin{equation}
\label{eq8}
\mathcal{L} = \mathcal{L}_{t}+\lambda\mathcal{L}_{f}
\end{equation}
\vspace{-0.1cm}
In detail, target frame loss and full sequence loss are:
\begin{equation}
\label{eq81}
\mathcal{L}_{t}=\frac{1}{J}\sum_{i=1}^{J}\left\|p_{i}-p_{i}^{gt} \right\|
\end{equation}
\vspace{-4pt}
\begin{equation}
\label{eq82}
\mathcal{L}_{f}=\frac{1}{T}\frac{1}{J}\sum_{t=1}^{T}\sum_{i=1}^{J}\left\|p_{t,i}-p_{t,i}^{gt} \right\|
\end{equation}
Where T and J refer to the sequence length and the number of joints. $p$ and $p^{gt}$ denote the predicted and ground truth 3D joints. $\lambda$ is the balance factor.
\vspace{-0.3cm}
\section{Experiments}
\subsection{Datasets}
The proposed architecture is evaluated on three 3D HPE benchmarks, i.e., Human3.6M \cite{ionescu2013human3}, MPI-INF-3DHP \cite{mehta2017monocular}, and HumanEva \cite{sigal2010humaneva}. Detailed descriptions and evluation matrics on these datasets are presented in Supplementary.

\subsection{Implementation Details}
The smaller, standard and larger G-SFormer are presented as G-SFormer-S, G-SFormer and G-SFormer-L on Human3.6M, with encoder-layer ($L1$), decoder-layer ($L2$) set as (3, 5), (4, 5), (8, 5) for 243 frames input. Residual connection across encoder layers is conducted only for G-SFormer-L. For G-SFormer-S with 81 frames and 27 frames input on MPI-INF-3DHP and Human-Eva, ($L1$, $L2$) is set as (3, 4) and (3, 3), respectively. We adopt the 2D pose detected by CPN \cite{chen2018cascaded} on Human3.6M following \cite{zhao2023poseformerv2,einfalt2023uplift,pavllo20193d,cheng20203d} and ground truth data on MPI-INF-3DHP and Human-Eva following \cite{zheng20213d,zhang2022mixste}. Models are trained from scratch on all these datasets. More detailed experimental settings are described in Supplementary.

\subsection{Comparison with State of the Arts}
\quad \ \textbf{Results on Human3.6M} We compare the proposed G-SFormer with SOTA approaches on Human3.6M dataset in Table \ref{tab:tab0}. Performance among 15 action categories under Protocol \#1 and Protocol \#2 metrics are reported. G-SFormer-L obtains average MPJPE of 41.7mm and average P-MPJPE of 33.4mm. With the implementation of the refinement module as \cite{li2022exploiting, einfalt2023uplift, shan2022p}, average results are further improved to 40.7mm and 33.0mm, respectively. G-SFormer series models outperform all the previous state-of-the-arts under Protocol \#1 and achieve the best MPJPE in 10 out of 15 actions. 


Since G-SFormer is proposed to realise efficient 3D HPE, the comprehensive properties of model size, speed and accuracy are the focus of assessment. Table \ref{tab2} presents properties in Paramter number, FLOPs, and MPJPE of G-SFormers and competitors. For a fair comparison, we constrain computational complexity within a comparable range. G-SFormer-S models show better performance than MixSTE \cite{zhang2022mixste} with only 13\%-14.9\% parameters and far less computational cost. It is also worth mentioning that compared with PoseFormer-V2 \cite{zhao2023poseformerv2} which realises high speed-accuracy trade-off among competitors, G-SFormer-S models outperform it by 1.9 mm-2.5 mm lower MPJPE, occupied with similar FLOPs and only 30.4\%-35\% parameters. Meanwhile, no pre-training stage deployed in transformer-based methods \cite{shan2022p, einfalt2023uplift} is required in our training pipeline, further confirming the advantages of G-SFormer.


\begin{table}
\scriptsize
\centering
\caption{Quantitative comparisons with SOTA methods on Human3.6M of MPJPE (mm) under Protocol \#1 and P-MPJPE (mm) under Protocol \#2, using CPN detected 2D poses as input. (*) uses the refinement module from \cite{cai2019exploiting}, (\dag) is the transformer-based methods. Best: \textbf{bold}, second best: \underline{underlined}.}
\label{tab:tab0}
\resizebox{\linewidth}{!}{ 
\begin{tabular}{ll|lllllllllllllll|l}
\hline
\multicolumn{2}{l|}{Protocol \#1} & Dir. & Disc. & Eat  & Greet & Phone & Photo & Pose & Pur. & Sit  & SitD. & Smoke & Wait & WalkD. & Walk & WalkT. & Avg. \\ \hline
Pavllo et al. \cite{pavllo20193d} (T =243)       & CVPR'19   & 45.2 & 46.7  & 43.3 & 45.6  & 48.1  & 55.1  & 44.6 & 44.3 & 57.3 & 65.8  & 47.1  & 44.0 & 49.0   & 32.8 & 33.9   & 46.8 \\
Cai et al. \cite{cai2019exploiting} (T=7)(*)            & ICCV'19   & 44.6 & 47.4  & 45.6 & 48.8  & 50.8  & 59.0  & 47.2 & 43.9 & 57.9 & 61.9  & 49.7  & 46.6 & 51.3   & 37.1 & 39.4   & 48.8 \\
Liu et al. \cite{liu2020attention} (T=243)          & CVPR'20   & 41.8 & 44.8  & 41.1 & 44.9  & 47.4  & 54.1  & 43.4 & 42.2 & 56.2 & 63.6  & 45.3  & 43.5 & 45.3   & 31.3 & 32.2   & 45.1 \\
UGCN \cite{wang2020motion} (T=96)                 & ECCV'20   & 40.2 & 42.5  & 42.6 & 41.1  & 46.7  & 56.7  & 41.4 & 42.3 & 56.2 & 60.4  & 46.3  & 42.2 & 46.2   & 31.7 & 31.0   & 44.5 \\
Chen et al. \cite{chen2021anatomy} (T=243)         & TCSVT'21  & 41.4 & 43.5  & 40.1 & 42.9  & 46.6  & 51.9  & 41.7 & 42.3 & 53.9 & 60.2  & 45.4  & 41.7 & 46.0   & 31.5 & 32.7   & 44.1 \\
PoseFormer \cite{zheng20213d} (T=81)(\dag)           & ICCV'21   & 41.5 & 44.8  & 39.8 & 42.5  & 46.5  & 51.6  & 42.1 & 42.0 & 53.3 & 60.7  & 45.5  & 43.3 & 46.1   & 31.8 & 32.2   & 44.3 \\
MHFormer \cite{li2022mhformer} (T=351)(\dag)             & CVPR'22   & 39.2 & 43.1  & 40.1 & 40.9  & 44.9  & 51.2  & 40.6 & 41.3 & 53.5 & 60.3  & 43.7  & 41.1 & 43.8   & 29.8 & 30.6   & 43.0 \\
Li et al. \cite{li2022exploiting} (T=351)(*\dag)  & TMM'22    & 40.3 & 43.3  & 40.2 & 42.3  & 45.6  & 52.3  & 41.8 & 40.5 & 55.9 & 60.6  & 44.2  & 43.0 & 44.2   & 30.0 & 30.2   & 43.7 \\
P-STMO \cite{shan2022p} (T=243)(*\dag)               & ECCV'22   & 38.4 & 42.1  & 39.8 & 40.2  & 45.2  & 48.9  & 40.4 & \textbf{38.3} & 53.8 & 57.3  & 43.9  & 41.6 & 42.2   & 29.3 & 29.3   & 42.1 \\
MixSTE \cite{zhang2022mixste} (T=243\dag)               & CVPR'22   & \underline{37.6} & \underline{40.9}  & \textbf{37.3} & 39.7  & \underline{42.3}  & 49.9  & 40.1 & 39.8 & \textbf{51.7} & \textbf{55.0}  & 42.1  & 39.8 &\underline{41.0}   & \textbf{27.9} & \textbf{27.9}   & \underline{40.9} \\

PoseFormerV2 \cite{zhao2023poseformerv2}(\dag)       & CVPR'23 &-&-&-&-&-&-&-&-&-&-&-&-&-&-&-&45.2\\
GLA-GCN \cite{yu2023gla} (T=243)             & ICCV'23   & 41.3 & 44.3  & 40.8 & 41.8  & 45.9  & 54.1  & 42.1 & 41.5 & 57.8 & 62.9  & 45.0  & 42.8 & 45.9   & 29.4 & 29.9   & 44.4 \\
Einfalt et al. \cite{einfalt2023uplift} (T=351)(*\dag)      & WACV'23   &39.6 &43.8 &40.2 &42.4 &46.5 &53.9 &42.3 &42.5 &55.7 &62.3 &45.1 &43.0 &44.7 &30.1 &30.8 &44.2 \\

\hline
\rowcolor[HTML]{DADADA} 
G-SFormer-S (T=243)(\dag)        & Ours          &40.3     & 43.2     &   39.6   & 40.8     & 43.9     &50.1       & 41.6     &40.1      & 53.1     &60.0      &43.3      &41.1      &43.4       & 29.8    &30.0        & 42.7     \\

\rowcolor[HTML]{DADADA}  
 G-SFormer-S (T=243)(*\dag)          &Ours          &38.3     & 41.6     &38.8     & 39.7      &  43.0     &49.8      &\underline{39.6}     & 39.6     & 52.3     & 57.9      & 42.3      &  40.0    & 42.0       &29.2     & 29.6      & 41.6     \\
 \rowcolor[HTML]{DADADA}  
 G-SFormer (T=243)(\dag)          &Ours          &39.8     & 42.9     &39.2     & 40.2      & 43.3     &49.9      &41.2     & 39.6     & 53.0     & 59.9      & 43.0      &  40.3    & 42.6       &29.4     & 29.8      &42.3   \\
 \rowcolor[HTML]{DADADA}  
  G-SFormer (T=243)(*\dag)         & Ours         &38.0     & 41.4     &\underline{37.9}     &\underline{39.3}     &42.8     &49.2      &\underline{39.6}     &  39.5     & \textbf{51.7}     &57.5     & \underline{42.0}      & \underline{39.5}   & 41.8       &29.0     & 29.6     &41.2   \\
  \rowcolor[HTML]{DADADA}   G-SFormer-L (T=243)(\dag)          &Ours          &39.3     & 42.2     &39.2     & 40.1      & 43.4     &\underline{48.5}      &41.0     & 39.2     & 52.7     & 57.3      & 42.5      &  39.9    & 41.8      &29.4     & 29.7      &41.7   \\
  \rowcolor[HTML]{DADADA}    G-SFormer-L (T=243)(*\dag)         & Ours         &\textbf{37.5}     & \textbf{40.5}     &38.7     & \textbf{39.0}      &\textbf{42.2}     &\textbf{47.4}      &\textbf{39.5}     &  \underline{38.9}     & \underline{52.0}     &\underline{55.4}     & \textbf{41.4}      & \textbf{39.3}    & \textbf{40.9}       &\underline{28.9}     & \underline{29.0}     &\textbf{40.7}   \\
\hline
\hline
\multicolumn{2}{l|}{Protocol \#2}  & Dir. & Disc. & Eat  & Greet & Phone & Photo & Pose & Pur. & Sit  & SitD. & Smoke & Wait & WalkD. & Walk & WalkT. & Avg.   \\
\hline
PoseFormer \cite{zheng20213d} (T=81)(\dag)      & ICCV'21   & 32.5 & 34.8  & 32.6 & 34.6  & 35.3  & 39.5  & 32.1 & 32.0 & 42.8 & 48.5  & 34.8  & 32.4 & 35.3   & 24.5 & 26.0   & 34.6 \\
Li et al. \cite{li2022exploiting} (T=351)(*\dag) & TMM'22    & 32.7 & 35.5  & 32.5 & 35.4  & 35.9  & 41.6  & 33.0 & 31.9 & 45.1 & 50.1  & 36.3  & 33.5 & 35.1   & 23.9 & 25.0   & 35.2 \\
P-STMO \cite{shan2022p} (T=243)(*\dag)               & ECCV'22   & \underline{31.3} & 35.2  & 32.9 & 33.9  & 35.4  & 39.3  & 32.5 & 31.5 & 44.6 & 48.2  & 36.3  & 32.9 & 34.4   & 23.8 & \underline{23.9}   & 34.4 \\
MixSTE \cite{zhang2022mixste} (T=243)(\dag)               & CVPR'22   & \textbf{30.8} & \textbf{33.1}  & \textbf{30.3} & \textbf{31.8}  & \textbf{33.1}  &39.1  & \textbf{31.1} & \textbf{30.5} & 42.5 & \textbf{44.5}  & \textbf{34.0}  & \textbf{30.8} & \textbf{32.7}   & \textbf{22.1} & \textbf{22.9}   & \textbf{32.6} \\

PoseFormerV2 \cite{zhao2023poseformerv2}(\dag)       & CVPR'23 &-&-&-&-&-&-&-&-&-&-&-&-&-&-&-&35.6\\
GLA-GCN \cite{yu2023gla} (T=243)   & ICCV'23 &32.4 &35.3 &32.6 &34.2 &35.0 &42.1 &32.1 &31.9 &45.5 &49.5 &36.1 &32.4 &35.6 &23.5 &24.7 &34.8 \\
Einfalt et al. \cite{einfalt2023uplift} (T=351)(*\dag)    & WACV'23  &32.7 &36.1 &33.4 &36.0 &36.1 &42.0 &33.3 &33.1 &45.4 &50.7 &37.0 &34.1 &35.9 &24.4 &25.4 &35.7\\

\hline
  \rowcolor[HTML]{DADADA}
 G-SFormer (T=243)(\dag)    &Ours & 31.9      &34.6  &32.6  &33.6    &33.9   &39.3  &32.2    &31.0  &42.9  &47.6  & 35.1 &32.1  &34.2  &24.0  &24.3      &33.9 \\
 \rowcolor[HTML]{DADADA}
 G-SFormer (T=243)(*\dag)    &Ours & 31.7      &34.2  &\underline{31.3}  &33.1    &33.7   &38.9  &\underline{31.3}    &31.4   &\underline{42.3}  &47.2  & 34.6 &31.7  &34.1  &23.8  &24.5      &33.6 \\
  \rowcolor[HTML]{DADADA}
 G-SFormer-L (T=243)(\dag)    &Ours & \underline{31.3}      &34.1  &32.3  &33.1    &33.7   &\underline{38.5} &32.0    &30.6   &42.5  &\underline{45.7}  &34.9 &31.4  &33.3  &24.0  &24.5      &33.4 \\
  \rowcolor[HTML]{DADADA}
 G-SFormer-L (T=243)(*\dag)    &Ours & \textbf{30.8}      &\underline{33.4}  &31.5  &\underline{32.5}    &\underline{33.3}   &\textbf{38.2}  &\textbf{31.1}    &\textbf{30.5}   &\textbf{41.9}  &46.1  & \underline{34.1} &\underline{31.1}  &\underline{32.9}  &\underline{23.3}  &24.1      &\underline{33.0} \\
\hline
\end{tabular}}
\end{table}

\begin{table}[]
\centering
\scriptsize
\caption{Quantitative comparisons with SOTA methods on Human3.6M under Parameter number, FLOPs, and MPJPE (mm). (*) indicates models with additional pre-training stage. Best: \textbf{bold}, second best: \underline{underlined}.}
\label{tab2}
\resizebox{0.54\linewidth}{!}{ 
\begin{tabular}{ll|l|l|l|l}
\hline
\multicolumn{2}{l|}{Method}      
& Frames &Params (M)  & FLOPs (M)  & MPJPE$\downarrow$ \\
\hline
PoseFormer \cite{zheng20213d}   & ICCV'21                      & 27 &9.59 & 452   & 47.0  \\
PoseFormer \cite{zheng20213d}   & ICCV'21                      & 81 &9.60 & 1358   & 44.3  \\
MHFormer \cite{li2022mhformer}  & CVPR'22                      & 27 &18.92 & 1030   & 45.9  \\
MHFormer \cite{li2022mhformer}  & CVPR'22                      & 81 &19.70  & 3132   & 44.5  \\
Li et al. \cite{li2022exploiting} & TMM'22                       & 81 &4.06 & 392    & 45.4  \\
Li et al. \cite{li2022exploiting} & TMM'22                       & 243 &4.23 & 1372   & 44.0  \\
Li et al. \cite{li2022exploiting} & TMM'22                       & 351 &4.34 & 2142   & 43.7  \\
P-STMO-S \cite{shan2022p} (*)          & ECCV'22                      & 81 &5.4 & 493    & 44.1  \\
P-STMO \cite{shan2022p} (*)             & ECCV'22                      & 243 &6.7 & 1737   & 42.8 \\
MixSTE \cite{zhang2022mixste}    & CVPR'22     & 27  & 33.65  & 30897 & 45.1  \\
MixSTE \cite{zhang2022mixste}   & CVPR'22     & 81  & 33.65  & 92692 & 42.7\\ 
Einfalt et al. \cite{einfalt2023uplift} (*)       & WACV'23       & 81 &10.36  & 543    & 45.5  \\
Einfalt et al. \cite{einfalt2023uplift} (*)       & WACV'23      & 351 &10.39  & 966    & 45.0  \\
PoseFormerV2 \cite{zhao2023poseformerv2}       & CVPR'23    & 81 &14.35  & 352     &46.0  \\
PoseFormerV2 \cite{zhao2023poseformerv2}       & CVPR'23    & 243   &14.35   & 1055   & 45.2  \\
\hline
\rowcolor[HTML]{DADADA} 
G-SFormer-S             & Ours                         & 81  &4.37   & 361    & 44.1  \\
\rowcolor[HTML]{DADADA} 
G-SFormer-S             & \cellcolor[HTML]{DADADA}Ours & 243 &5.02  & 1092    & 42.7  \\
\rowcolor[HTML]{DADADA} 
G-SFormer               & \cellcolor[HTML]{DADADA}Ours & 243 &5.54  & 1346    & \underline{42.3}   \\
\rowcolor[HTML]{DADADA} 
G-SFormer-L               & \cellcolor[HTML]{DADADA}Ours & 243 &7.65  & 2366    & \textbf{41.7}   \\
\hline
\end{tabular}}
\end{table}

\begin{figure}
\begin{minipage}[t]{0.42\textwidth}
\makeatletter\def\@captype{table}
\captionsetup{font={footnotesize}}
\caption{Quantitative comparisons with SOTA methods on MPI-INF-3DHP dataset. Best: \textbf{bold}, second best: \underline{underlined}.}
\label{tab3}
\centering
\resizebox{\linewidth}{!}{ 
\begin{tabular}{ll|lll}
\hline
\multicolumn{2}{l|}{Method}         & PCK$\uparrow$   & AUC$\uparrow$   & MPJPE$\downarrow$ \\ \hline
Pavllo et al. \cite{pavllo20193d} (T=81)        &CVPR’19  & 86.0  & 51.9  & 84.0  \\
UGCN \cite{wang2020motion} (T=96)               &ECCV’20  & 86.9  & 62.1  & 68.1  \\
Hu et al. \cite{hu2021conditional} (T=96)       &MM’21  & 97.9  & 69.5  & 42.5  \\
Chen et al. \cite{chen2021anatomy} (T=81)       &TCSVT’21  & 87.9  & 54.0  & 78.8  \\
PoseFormer \cite{zheng20213d} (T=9)             &ICCV’21  & 88.6  & 56.4  & 77.1  \\
P-STMO \cite{shan2022p} (T=81)                  &ECCV’22  & 97.9  & 75.8  & 32.2  \\
MixSTE \cite{zhang2022mixste} (T=27)            &CVPR’22  & 94.4  & 66.5  & 54.9  \\
Einfalt et al. \cite{einfalt2023uplift} (T=81)  &WACV’23  & 95.4  & 67.6  & 46.9  \\
GLA-GCN \cite{yu2023gla} (T=81)                 &ICCV’23  & \textbf{98.5} & \underline{79.1} & 27.8 \\
PoseFormerV2 \cite{zhao2023poseformerv2} (T=81) &CVPR’23  & 97.9  & 78.8  & 27.8  \\ \hline
\rowcolor[HTML]{DADADA} 
G-SFormer-S (T=27)  &Ours  &98.0   &79.0     &\underline{27.7} \\
\rowcolor[HTML]{DADADA} 
G-SFormer-S (T=81)                &Ours  & \underline{98.4}  & \textbf{80.2}  & \textbf{25.7} \\
\hline
\end{tabular}}
\end{minipage}
\quad
\begin{minipage}[t]{0.47\textwidth}
\makeatletter\def\@captype{table}
\captionsetup{font={footnotesize}}
\caption{Quantitative comparisons with SOTA methods on Human-Eva dataset of MPJPE (mm) under Protocol \#1. (FT) indicates models finetuned from the Human3.6M pre-training models and (†) is our re-implementation results.}
\label{tab:tab4}
\centering
\resizebox{\linewidth}{!}{ 
\begin{tabular}{ll|lll|lll|l}
\hline
\multicolumn{2}{c|}{}     & \multicolumn{3}{c|}{Walk}         & \multicolumn{3}{c|}{Jog}                                                        & \multicolumn{1}{c}{}                      \\
\multicolumn{2}{c|}{\textbf{Protocol \#1}} & \multicolumn{1}{c}{S1}   & \multicolumn{1}{c}{S2}   & \multicolumn{1}{c|}{S3}   & \multicolumn{1}{c}{S1}   & \multicolumn{1}{c}{S2}   & \multicolumn{1}{c|}{S3}   & \multicolumn{1}{c}{Avg.} \\ \hline
PoseFormer  \cite{zheng20213d} (T=43)                     &                      & 16.3 & 11.0 & 47.1 & 25.0 & 15.2 & 15.1 & 21.6                                      \\
PoseFormer  \cite{zheng20213d} (T=43, FT)                 &         & 14.4 & 10.2 & 46.6 & 22.7 &13.4 & 13.4 & 20.1      \\
MixSTE \cite{zhang2022mixste} (T=43)                         &                      & 16.2                     & 14.2                     & 21.6                      & 24.6                     & 23.2                     & 25.8                      & 20.9                                      \\
MixSTE \cite{zhang2022mixste} (T=43, FT)                     &                      & \textbf{12.7}                     & 10.9                     & \textbf{17.6}                      & 22.6                     & 15.8                     & 17.0                      & 16.1                                      \\
PoseFormer-V2 \cite{zhao2023poseformerv2} (T=81) (\dag)                &                      & 18.3                     & 12.9                     & 35.1                      & 28.9                     & 16.4                     & 17.7                      & 21.5                                      \\ \hline
\rowcolor[HTML]{DADADA} 
G-SFormer-S (T=81)                           &                 & 13.2                     & \textbf{9.3}                      & 25.9                      & \textbf{20.7}                     & \textbf{11.3}                     & \textbf{12.7}                      & \textbf{15.5}                                      \\ \hline
\end{tabular}}
\end{minipage}
\end{figure}

\textbf{Results on MPI-INF-3DHP}  Table \ref{tab3} reports comparison with SOTA methods on MPI-INF-3DHP. G-SFormer-S achieves the best AUC and MPJPE that outperforms the most competitive methods by 1.1\% and 2.1mm, respectively. It also reaches the second best PCK with the gap of only 0.1\%, but with only 34.7\% computational cost compared with GLA-GCN \cite{yu2023gla} (361M vs. 1039M FLOPs). It is also worth mentioning that with 27 frames input, G-SFormer-S outperforms MixSTE \cite{zhang2022mixste} by a large margin of 3.6\%, 12.5\% and 27.2mm in PCK, AUC and MPJPE, occupying only 0.39\% computational cost (120M vs. 30897M FLOPs) and 11\% parameters (3.71M vs. 33.65M). In this more challenging dataset containing complex outdoor poses, the substantial performance enhancement proves superiority and stability of our method.

\textbf{Results on HumanEva} To further explore the generalization of our method, we perform experiments on HumanEva and present results in Table \ref{tab:tab4}. Since HumanEva is a small dataset, prior methods usually pre-train model on large-scale datasets \cite{zheng20213d, zhang2022mixste}. G-SFormer-S obtains the best result without any pre-training operation, reflecting strong generalization ability towards dataset in various scales.

\textbf{Qualitative Results} 

\begin{figure}
\centering
\begin{minipage}[h]{0.48\textwidth}
\centering
\includegraphics[width=\textwidth]{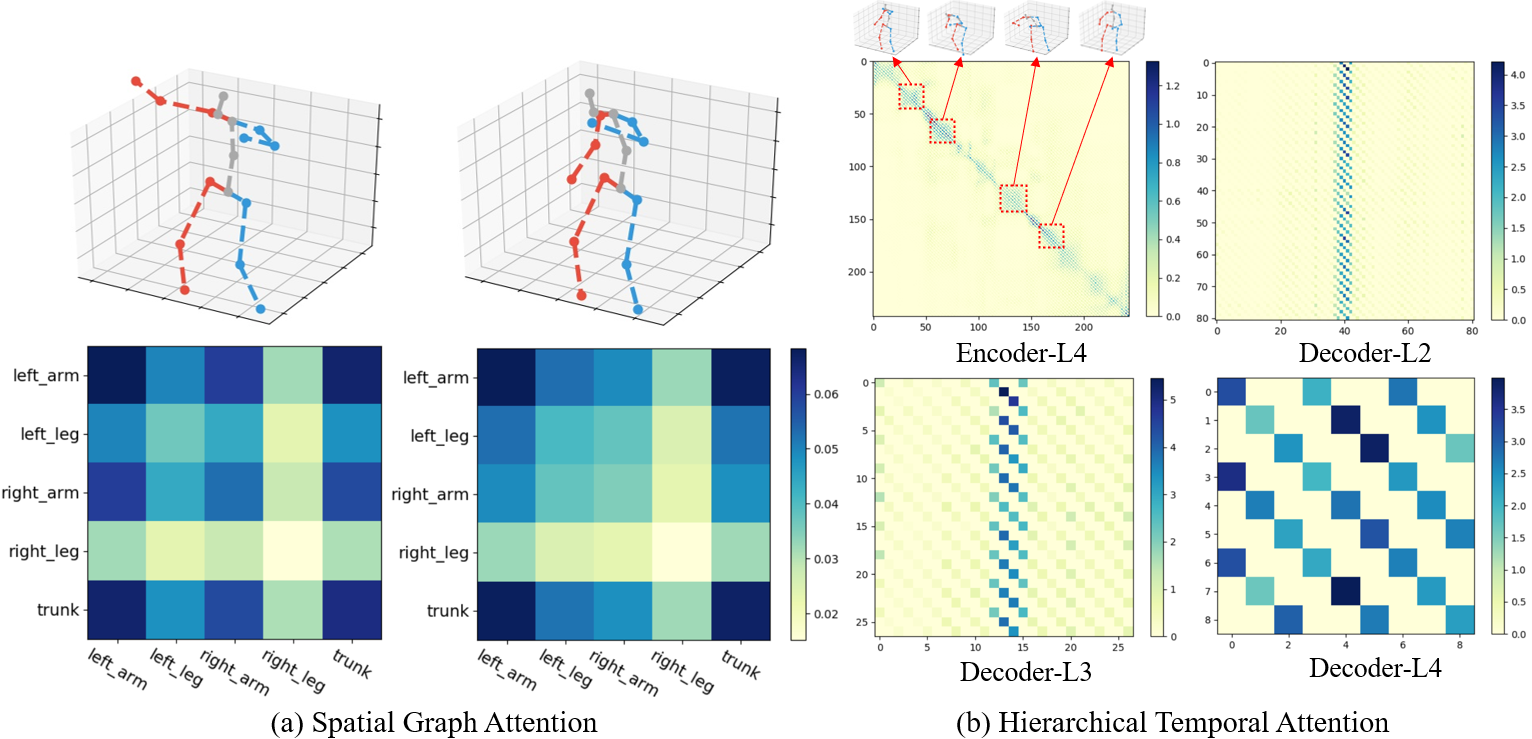}
\captionsetup{font={footnotesize}}
\caption{Visualized attention maps of "Greeting" action in (a) Spatial Graph Encoder and (b) Skipped Transformer based Temporal Encoder/Decoder. Attention maps corresponding to multiple heads of Skipped Transformer are summed to obtain the comprehensive temporal correlation distribution. }
\label{fig5}
\end{minipage}
\quad
\begin{minipage}{0.48\textwidth}
\centering
\includegraphics[width=\textwidth]{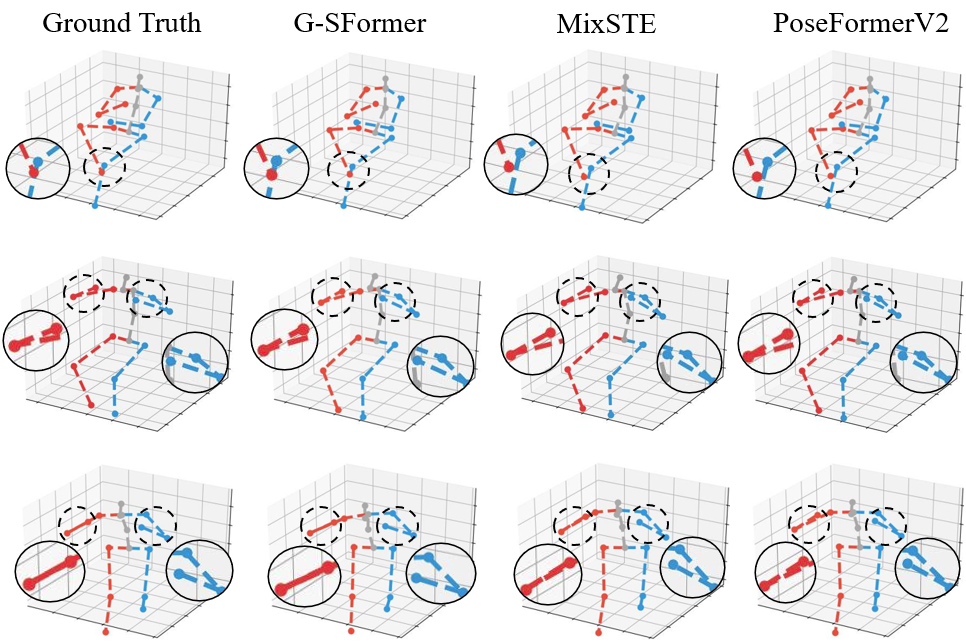}
\captionsetup{font={footnotesize}}
\caption{Visualized qualitative comparison with SOTA methods of MixSTE \cite{zhang2022mixste} and PoseFormerV2 \cite{zhao2023poseformerv2} on Human3.6M dataset. }
\label{fig7}
\end{minipage}
\end{figure}

\textbf{Attention Visualization.} Figure \ref{fig5} presents spatial and temporal attention distributions. Greeting action of S11 subject on Human3.6M is applied for visualization. It can be seen from (a) that attention concentrates on left arm, right arm and trunk which are main parts for hugging motion. In the temporal domain (b), we can see that global attention is built at full-sequence scale from the attention map of Encoder-L4 layer, and the strength increases during the periods when typical hugging action is performed. As the hierarchical decoding stage progresses, temporal tokens are established with dependencies with target frame and its neighbours in shallow Decoder layers (e.g., L2 with 81 tokens and L3 with 27 tokens). Thereafter, global range dependency is built since tokens in deeper decoder layers (e.g., L4 with 9 tokens) have stronger global representation ability. \textit{It is worth noting that stronger attention weights distributed sparsely in both local and global temporal intervals.} Compared with the dense and clustered attention maps of conventional transformer-based methods \cite{zheng20213d, shan2022p}, the redundant connections is greatly reduced and global-range dependencies is established among more distinct and representative tokens. Specific comparisons are presented in Supplementary Material.


\textbf{Visualized Comparison.} We also present the visualized results of G-SFormer and SOTA methods \cite{zhang2022mixste, zhao2023poseformerv2} in and out of the datasets. Figure \ref{fig7} shows the comparisons in actions of Smoking, Sitting and Photoing from Human3.6M dataset. G-SFormer realise more accurate 3D pose reconstruction across all three challenging examples with complicated body posture.

\begin{figure}
    \centering
    \includegraphics[width=1\textwidth]{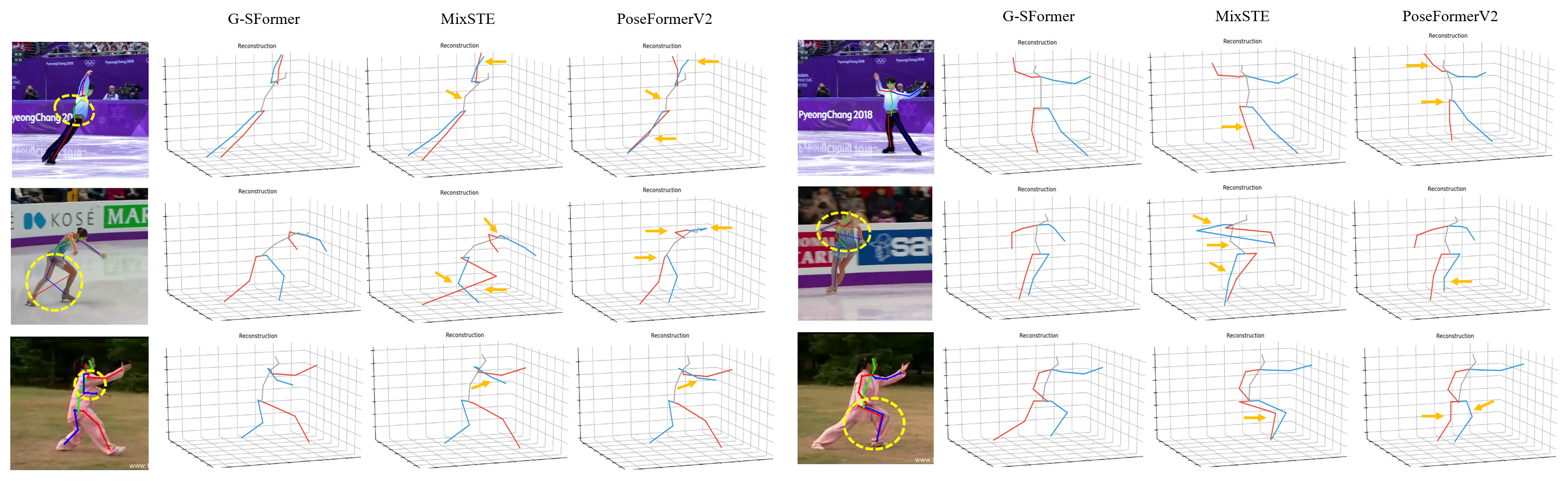}
    \caption{Visualized comparison with SOTA methods in wild videos. The erroneously detected 2D joints are marked in yellow circles and the inaccurately constructed 3D joint locations are marked with arrows. G-SFormer can still estimate correct 3D poses in such hard cases.}
    \label{fig6}
\end{figure}

Videos in-the-wild are even more challenging for 3D HPE task due to complex and changeable movements. Factors such as self-occlusion, fast-motion, as well as unconstrained environment and image quality significantly increase the error rate of detected joint coordinates\cite{chen2018cascaded,sun2019deep}. Typical hard cases are presented in Figure \ref{fig6}, including detection errors such as joint position deviation, left-right switch, confusion caused by self-occlusion and miss/coincidence detection, all highlighted by yellow circles. G-SFormer estimates more refined and structurally reasonable 3D poses than existing methods, demonstrating its strong temporal modeling capacity by incorporating contextual information to supplement the deficiency in current frame. In contrast, MixSTE \cite{zhang2022mixste} exhibits higher sensitivity to noise, attributed to its excessive focus on local information while lacking a comprehensive understanding of global motion process of the pose sequence. Although PoseFormerV2 \cite{zhao2023poseformerv2} extracts global-view representation by low-frequency DCT coefficients, this compression manner discards features in temporal domain and inevitably causes a loss of accuracy. G-SFormer provides an efficient and robust approach to establish global-range connections without information loss. 
\vspace{-0.5cm}



\vspace{-1.5cm}

\vspace{-1cm}
\subsection{Ablation Study}
To verify the effectiveness main proposals, extensive ablation studies are conducted on Human3.6M. The presented analysis is based on G-SFormer-S with 243 frames input.

\begin{figure}
    \centering
    \includegraphics[width=0.3\textwidth]{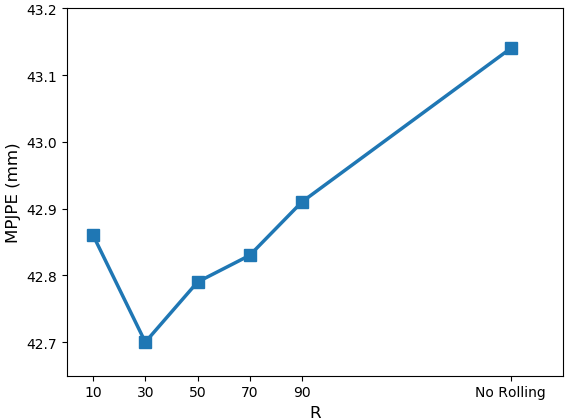}
    \caption{MPJPE (mm) of G-SFormer-S trained and tested with different Data Rolling threshold $R$.}
    \label{fig8}
\end{figure}

\textbf{Data Completion Methods} Figure \ref{fig8} shows the variation of MPJPE as the Data Rolling threshold R changes. Lowest error occurs when R is with the value of 30, which introduces more than 0.4 mm error drop compared with edge padding strategy. On the other hand, the impact of Data Expanding is not obvious, which is less than 0.1mm. To account for datasets of different scales, R is typically set to a ratio of 10\% - 20\% relative to various input lengths.

\begin{table}[]
\scriptsize
\centering
\caption{Ablation studies of the impact of different components and skipped factor $m$ of G-SFormer-S. Experiments are conducted on Human3.6M dataset of MPJPE (mm).}
\label{tab:tab5}
\setlength\tabcolsep{5pt}
\begin{tabular}{ll|llll|ll}
\hline
  \multicolumn{2}{l|}{No.}     & m &SSA &Parts Enco & Spatial Attn & FLOPs (M) & MPJPE \\
                \hline
1 &Spatial-MLP     & 3 &  \ding{52} &\ding{52}  &\ding{56}     & 1100         & 43.6      \\
2 &Joint-wise GCN     & 3 &  \ding{52} &\ding{56}  &\ding{52}     & 1387    & 44.4     \\
3 &VT-Strided Conv         & 1 & \ding{56}   &\ding{52}   & \ding{52}     &1219  & 43.7      \\
4 &VT-Conv & 1 & \ding{56}     &\ding{52}        & \ding{52}     &2111          & 44.0      \\
5 &G-SFormer-S-m3  & 3 & \ding{52}     &\ding{52}        &\ding{52}     & 1092         & \textbf{42.7}      \\
6 &G-SFormer-S-m5  & 5 & \ding{52}    &\ding{52}        &\ding{52}     & 1038         & 43.0       \\
7 &G-SFormer-S-m7  & 7 & \ding{52}       &\ding{52}      &\ding{52}     & 1016         & 43.8     \\
8 &G-SFormer-S-m9  & 9 & \ding{52}     &\ding{52}        &\ding{52}     & 997         & 43.9  \\   
\hline
\end{tabular}
\end{table}

\textbf{Impact of Components} Table \ref{tab:tab5} lists the impact of different components and various skipped factor $m$ to overall performance of G-SFormer-S. In row 1 and 2, MLP layers and Joint-wise GCN are used to replace Part-based Adaptive GNN, leading to performance drop of up to 1.7 mm. To verify the effectiveness of Skipped Transformer, Vanilla Transformer-based (VT) models are presented in row 2-3, incorporated Convolutional layer in \cite{zheng20213d, zhao2023poseformerv2} and Strided Convolutional layer in \cite{cai2019exploiting, shan2022p} for temporal aggregation, respectively. Performance drop 1.0-1.3 mm can be observed with \textit{up to nearly twice computational cost}. It is important to highlight that G-SFormer exhibits the highest performance without redundant spatial and temporal connections in Joint-wise GCN and VT series structures, further demonstrating its superiority in both speed and accuracy. To validate the effect of factor $m$ in Skipped Transformer, G-SFormer-S with $m=1,3,5,7,9$ is evaluated in row 4-8 ($m=1$ equals to VT-Conv). We draw the conclusion that $m$ should be restrained within an appropriate range, as a proper $m$ strengthens global dependencies and reduces computational complexity. However, excessively high value of $m$ will degrade the temporal coherence of pose sequence. Overall, the hyperparameter $m$ enables G-SFormer with an adaptive architecture for different speed and accuracy requirements.

\vspace{-0.5cm}

\section{Conclusion}
\vspace{-0.2cm}

In this paper, we present a simple yet effective G-SFormer architecture, exploiting the capabilities of Part-based Adaptive GNN and Skipped Transformer for lifting-based 3D HPE task. The totally data-driven adaptive GNN establishes body part correlations within each frame and the Skipped Transformer provides a concise and adaptive feature aggregation manner for full pose sequence. Experiments demonstrate that G-SFormer effectively reduces redundant connections and constructs informative global-range dependencies, realizing high-accuracy, efficient, and robust 3D pose estimation for both experimental and wild videos.

%
%
\bibliographystyle{splncs04}
\bibliography{main}
\end{document}